\documentclass{article}

\usepackage{microtype}
\usepackage{graphicx}
\usepackage{subfigure}
\usepackage{booktabs} 
\usepackage{url}
\usepackage{amsmath}
\usepackage{natbib}
\usepackage{algorithm}
\usepackage[noend]{algorithmic}
\usepackage{amsfonts}
\usepackage{color}
\usepackage{multirow}
\usepackage{makecell}
\usepackage[table]{xcolor}
\usepackage{comment}
\usepackage{wrapfig}
\usepackage{todonotes}

\usepackage{hyperref}

\usepackage[accepted]{mlsys2025}

\mlsystitlerunning{DSD: A Distributed Speculative Decoding Solution for Edge-Cloud Agile Large Model Serving}

\begin{document}

\twocolumn[
\mlsystitle{DSD: A Distributed Speculative Decoding Solution for Edge-Cloud Agile Large Model Serving}


\begin{mlsysauthorlist}
\mlsysauthor{Fengze Yu}{nyu}
\mlsysauthor{Leshu Li}{umn}
\mlsysauthor{Brad McDanel}{fandm}
\mlsysauthor{Sai Qian Zhang}{nyu}
\end{mlsysauthorlist}

\mlsysaffiliation{nyu}{New York University, New York, NY, USA}
\mlsysaffiliation{umn}{University of Minnesota Twin Cities, Minneapolis, MN, USA}
\mlsysaffiliation{fandm}{Franklin \& Marshall College, Lancaster, PA, USA}

\mlsyscorrespondingauthor{Sai Qian Zhang}{sai.zhang@nyu.edu}

\mlsyskeywords{Machine Learning, MLSys}

\vskip 0.3in

\begin{abstract}
Large language model (LLM) inference often suffers from high decoding latency and limited scalability across heterogeneous edge–cloud environments. Existing speculative decoding (SD) techniques accelerate token generation but remain confined to single-node execution. We propose \textit{DSD}, a distributed speculative decoding framework that extends SD to multi-device deployments through coordinated draft–target execution. Given the lack of prior work on simulating this paradigm, we first introduce \textit{DSD-Sim}, a discrete-event simulator that captures network, batching, and scheduling dynamics. Building on insights from DSD-Sim, we further design an \textit{Adaptive Window Control} (AWC) policy that dynamically adjusts speculation window size to optimize throughput. Experiments across diverse workloads show that DSD achieves up to \textbf{1.1$\times$} speedup and \textbf{9.7\%} higher throughput over existing SD baselines, enabling agile and scalable LLM serving across edge and cloud.
\end{abstract}
]

\printAffiliationsAndNotice{}

\section{Introduction}
\label{sec:introduction}
Large Language Models (LLMs) have become a cornerstone of modern artificial intelligence by advancing natural language understanding and generation. These models enable machines to process, interpret, and produce human-like text, supporting a wide range of applications across domains.
Despite their impressive capabilities, LLMs continue to suffer from significant processing latency, primarily due to the computationally intensive prefilling and autoregressive decoding stages. To address this, speculative decoding (SD)~\cite{stern2018blockwise, leviathan2023fast} has emerged as an effective acceleration strategy. SD decomposes the autoregressive generation process into two stages: a lightweight drafting stage that proposes multiple candidate tokens, followed by a verification stage in which the target LLM evaluates these candidates in parallel. This two-phase design improves decoding throughput while preserving output quality via a selective acceptance–rejection mechanism.

However, although SD can substantially accelerate LLM inference, it remains insufficient to meet the scalability requirements of large-scale or real-time deployments. SD primarily improves token generation efficiency within a single device or limited hardware environment. Yet, modern LLM applications—such as interactive AI chatbots~\cite{openai2023gpt4}, intelligent tutoring systems~\cite{kasneci2023chatgpt}, and AI recommendation systems~\cite{gao2023chat}—often need to serve thousands of concurrent users while maintaining low latency.

\begin{figure*}[t]
  \centering
  \includegraphics[width=2\columnwidth]{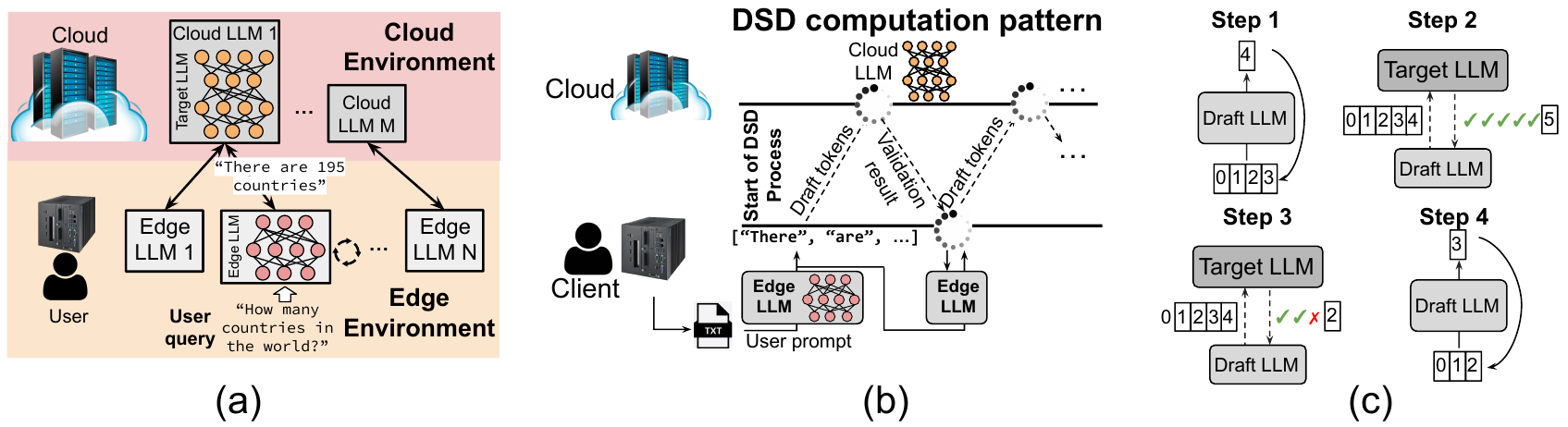}
  \caption{(a) Distributed edge–cloud environment for speculative decoding. (b) Joint processing between edge and cloud servers during SD operation. (c) Illustration of the speculative decoding workflow.}
  \label{fig:sd_intro}
\end{figure*}

Crucially, most existing SD frameworks focus on algorithmic efficiency and assume that the draft and target models reside on the same device~\cite{chen2023accelerating} or, at most, across two nodes~\cite{liu2024pearl}. As a result, they remain limited by the inherent sequential dependency between the drafting and verification stages, which constrains overall throughput. Moreover, the large computational footprint of the target LLM makes it impractical to deploy within resource-constrained edge environments, further restricting SD's applicability in real-world edge–cloud scenarios.

As model complexity and application demands escalate, achieving true scalability necessitates extending SD beyond traditional single-node or small-scale deployments. A distributed SD framework can leverage both model and data parallelism to offload and coordinate computation across edge and cloud resources, thereby reducing latency, improving resource utilization, and enabling efficient large-scale LLM inference for real-world applications.

An example deployment scenario is illustrated in Figure~\ref{fig:sd_intro} (a). Given the large scale of the target model, it is hosted across $M$ cloud servers responsible for performing the verification stage on behalf of $N$ edge LLMs, each serving as a draft model on resource-constrained edge devices. During operation, an edge LLM receives a user request and context prompt, generates a set of $\gamma$ draft tokens, and transmits them to the cloud LLM for verification. Based on the verification feedback, the edge LLM continues generating subsequent tokens until the completion of the response sequence. The process is described in Figure~\ref{fig:sd_intro} (b).

In this paper, we introduce~\textit{DSD}, a unified distributed speculative decoding framework designed to enable scalable LLM serving across multiple nodes. Our contributions can be summarized as follows:
\begin{itemize}
\item We develop \textit{DSD-Sim}, a large-scale DSD simulation framework that accurately models the performance behavior of distributed speculative decoding systems under diverse system and network conditions. DSD-Sim follows a modular design philosophy and is compatible with a wide range of existing simulation frameworks for LLM inference, networking, and distributed computing, allowing seamless integration and extensibility.
\item Motivated by the performance bottlenecks identified through DSD-Sim, we further introduce an~\textit{Adaptive Window Control} (AWC) mechanism to enhance the overall throughput of the DSD framework. AWC employs a deep learning–based approach that dynamically adjusts the window configuration based on real-time system and algorithmic conditions, enabling the system to maintain optimal performance across varying workloads and network environments.

\item Extensive experiments on GSM8K, CNN/DailyMail, and HumanEval benchmarks show that DSD combined with AWC consistently enhances system efficiency, achieving up to 9.7\% higher throughput and 11\% lower TPOT than fixed-threshold and heuristic scheduling baselines, all without manual tuning.
\end{itemize}

\section{Background and Related Work}
\label{sec:bg}

\subsection{Speculative Decoding}
\label{sec:bg:sd}
Speculative decoding is an effective technique for mitigating the sequential bottleneck in language model inference~\citep{stern2018blockwise, chen2023accelerating}. It decomposes the decoding process into two stages: a lightweight draft model quickly proposes a sequence of candidate tokens, which are then verified in parallel by a more accurate target model, as illustrated in Figure~\ref{fig:sd_intro} (c).

Let the draft model $M_{da}$ generate $\gamma$ draft tokens ($t_1, \dots, t_\gamma$) in each iteration (step 1 of Figure~\ref{fig:sd_intro} (c)), where $\gamma$ denotes the~\textit{speculation window size}. During verification, the target model $M_{ta}$ evaluates all draft tokens in parallel but accepts them sequentially. If all tokens are accepted, $M_{da}$ proceeds to produce the next batch of candidates (step 2 of Figure~\ref{fig:sd_intro} (c)). Otherwise, if a mismatch occurs at position $i$, all tokens from $t_i$ onward are discarded, and the target model’s own sampled token $t'_{i}$ is used instead (step 3 of Figure~\ref{fig:sd_intro} (c)). The resulting accepted sequence is therefore \( (t_1, \dots, t_{i-1}, t'_i) \), and the draft model will generate subsequent tokens based on the prefix determined by $M_{ta}$ (step 4 of Figure~\ref{fig:sd_intro} (c)).

Assuming a per-token acceptance probability $\alpha$, the expected number of accepted draft tokens is:
\begin{equation}
\mathbb{E}[\tau] = \frac{1 - \alpha^{\gamma + 1}}{1 - \alpha}.
\end{equation}
Let $c$ denote the cost ratio between the draft and target models per token. The expected speedup over standard decoding with $M_{ta}$ is then:
\begin{equation}
S = \frac{1 - \alpha^{\gamma + 1}}{(1 - \alpha)(c\gamma + 1)}.
\end{equation}
This acceptance mechanism is compatible with various sampling temperatures. By allowing parallel draft generation while maintaining output fidelity through sequential verification, speculative decoding effectively relaxes the step-by-step dependency inherent in traditional autoregressive decoding.

\subsection{Split Computing}
\label{sec:split_computing}
Split neural network computing has attracted substantial attention from both academia and industry, as reflected in a broad range of studies~\cite{hauswald2014hybrid, teerapittayanon2016branchynet, kang2017neurosurgeon, teerapittayanon2017distributed, karjee2022split, luo2023split, mubark2024asap, lee2023wireless, feltin2023dnn, zeng2020coedge, ding2023resource, kang2022dnn, matsubara2019distilled, matsubara2022split, lin2024murmuration, zhang2022multi, dong2022spherefed, zhang2024learning}. Both commercial and open-source systems have adopted split learning and distributed inference in practice~\cite{pysyft, splitnn}.

Among various partitioning strategies~\cite{kang2017neurosurgeon, zhang2020adaptive}, layer-wise partitioning is the most widely adopted~\cite{hauswald2014hybrid, teerapittayanon2016branchynet, kang2017neurosurgeon, teerapittayanon2017distributed}. In this approach, a deep neural network (DNN) is divided into two or more segments that are executed collaboratively across multiple devices.

One of the earliest works in this direction, Hauswald \emph{et al.}~\cite{hauswald2014hybrid}, proposed offloading the later stages of image classification computation to cloud servers. Following this, Neurosurgeon~\cite{kang2017neurosurgeon} and DDNN~\cite{teerapittayanon2017distributed} introduced automated frameworks to dynamically distribute DNN workloads between mobile devices and cloud servers, optimizing for factors such as network latency and energy efficiency. Meanwhile, BranchyNet~\cite{teerapittayanon2016branchynet} incorporated early-exit branches within DNN architectures, enabling adaptive inference based on input complexity to further reduce end-to-end processing latency. To the best of our knowledge, DSD represents the first split computing framework specifically designed for speculative decoding.

\subsection{Simulation Framework for Large Models}
\label{sec:bg:llm-sim}

As LLMs continue to increase in both complexity and deployment scale, their performance evaluation increasingly depends on simulation frameworks that capture the interplay among workload characteristics, system architecture, and resource constraints. These frameworks operate across multiple levels of abstraction, ranging from GPU microarchitecture to datacenter-scale distributed systems. In the following, we review representative simulators at each level.

\paragraph{GPU-Level Simulators.} At the hardware level, GPU simulators enable cycle-accurate modeling of GPU architectures for diverse workloads. GPGPU-Sim~\cite{gpgpusim} offers detailed simulation of modern NVIDIA GPUs executing CUDA applications, with support for Tensor Cores, dynamic parallelism, and integrated energy estimation via GPUWattch. Building upon this foundation, Accel-Sim~\cite{accelsim_isca2020} extends the framework with an extensible, trace-driven design that directly consumes native SASS instructions, substantially reducing the effort needed to model new GPU architectures while preserving validation accuracy within $20\%$ of real hardware measurements.

\paragraph{LLM Inference Simulators.} The emergence of large language model (LLM) serving systems has driven the development of specialized simulation frameworks that capture the unique dynamics of autoregressive inference. VIDUR~\cite{vidur_mlsys2024}, the first large-scale simulator for LLM inference, models operator performance using empirical profiling and predictive modeling, achieving an average latency estimation error of approximately 9\%. Building on this direction, LLMServingSim~\cite{llmservingsim_iiswc2024} advances hardware–software co-simulation by extending ASTRA-sim to jointly model LLM-specific accelerators and serving system software. It integrates GPU kernel simulation with system-level request scheduling and batching, providing up to 91.5$\times$ faster simulation than prior accelerator simulators while maintaining an error below 14.7\% compared to real GPU-based serving systems. Complementing these efforts, ReaLLM~\cite{reallm_asap2025} introduces a trace-driven framework with a built-in generator that leverages production workloads from the Azure LLM Inference Dataset. Through linear interpolation–based latency prediction for matrix multiplication, ReaLLM accelerates simulation while maintaining low error rates, enabling rapid evaluation of service-level objectives across diverse system configurations. DSD-Sim leverages the VIDUR framework as its foundation for modeling the performance of distributed SD, yet it maintains compatibility with all previously discussed LLM simulation frameworks.

\paragraph{Distributed Neural Network Simulators.} Numerous efforts have been made to simulate the behavior of distributed training and inference. ASTRA-sim~\cite{astrasim_ispass2020,astrasim2_ispass2023} models the end-to-end software–hardware stack, encompassing workload scheduling, collective communication algorithms, and hierarchical network topologies. Its second version extends this capability by supporting arbitrary model-parallel strategies through graph-based implementations and by providing analytical performance estimates for multi-dimensional heterogeneous topologies, thereby enabling scalable simulation-based exploration of distributed training platforms.

\paragraph{Advanced LLM Serving Frameworks.} Recent studies highlight the critical role of specialized architectures and scheduling policies in optimizing LLM serving. DistServe~\cite{distserve_osdi2024} disaggregates prefill and decoding stages across distinct GPUs to eliminate phase interference, achieving up to 7.4$\times$ higher goodput under stringent Service-Level Objectives (SLO) constraints. Sarathi~\cite{agrawal2023sarathi} proposes chunked prefill, which partitions long input prompts into fixed-size chunks to enable decode-maximal batching, improving decode throughput by as much as 10$\times$ for LLaMA-13B. Building on this, Sarathi-Serve~\cite{sarathi_osdi2024} introduces stall-free scheduling mechanisms to further enhance latency control during serving.

While existing frameworks effectively model single-node LLM inference or distributed training, they fall short in addressing the unique challenges of distributed SD. Unlike conventional serving, where each request is executed independently on a single device, distributed SD introduces cross-device dependencies between draft generation and target verification, with performance tightly coupled to network round-trip time, token acceptance rates, and dynamic speculation window size. Furthermore, the interaction among prefill queueing, speculation window size, and target model utilization leads to complex performance bottlenecks that cannot be captured by single-node simulators such as VIDUR or system-agnostic frameworks like ASTRA-sim. These limitations motivate the development of a specialized simulation framework.

\section{Simulation Framework for Distributed Speculative Decoding}
\begin{figure*}[t]
  \centering
  \includegraphics[width=2\columnwidth]{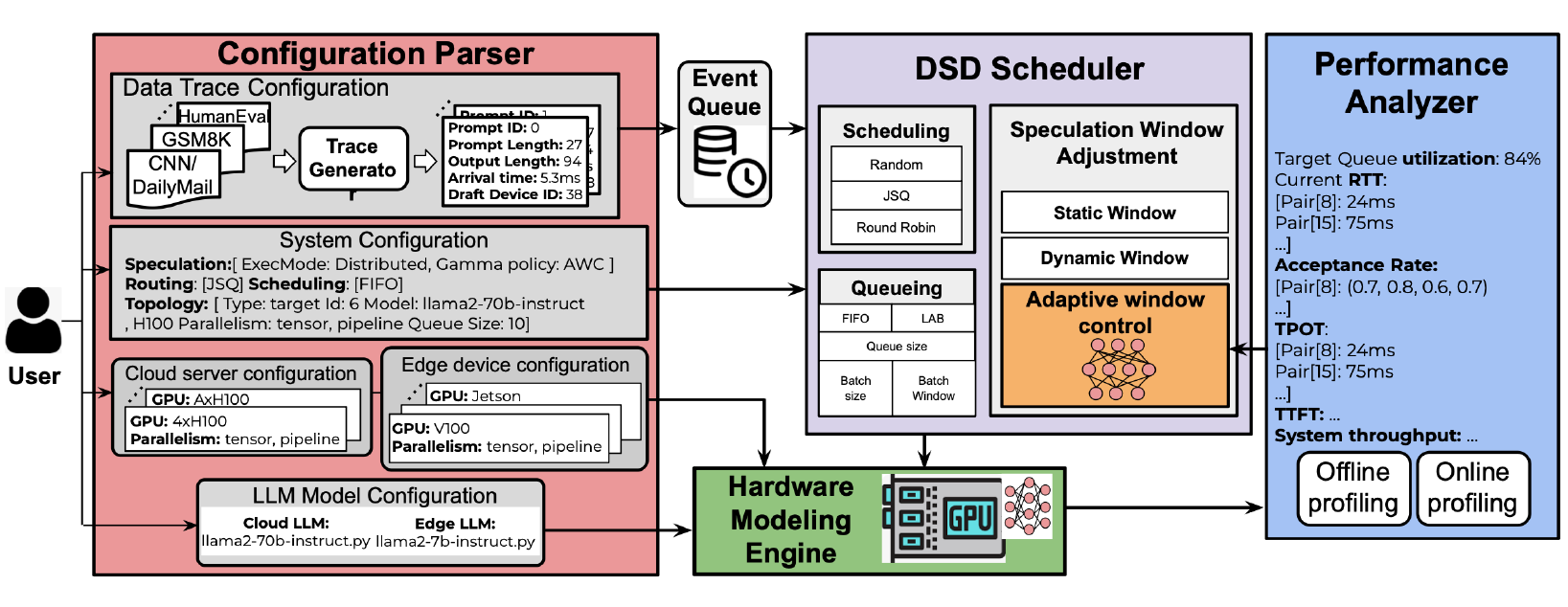}
  \caption{The configuration parser ingests YAML configuration files and workload traces, routing requests through the DSD scheduler. The scheduler then coordinates the hardware modeling engine to generate detailed system performance outputs, which are subsequently processed by the performance analyzer for SLO evaluation.}
  \label{fig:system-overview}
\end{figure*}

As discussed in Section~\ref{sec:bg:llm-sim}, due to the lack of an existing simulation framework for distributed SD, we introduce \textit{DSD-Sim}, a request-level discrete-event simulator tailored for distributed SD. Given a deployment description and workload traces, DSD-Sim models routing, batching, speculation/verification iterations, and network effects to produce per-request and system-level SLO metrics. Figure~\ref{fig:system-overview} summarizes the data flow.

\subsection{Overview}

As shown in Figure~\ref{fig:system-overview}, DSD-Sim comprises four primary components: a~\textit{configuration parser}, a core~\textit{DSD scheduler}, a~\textit{hardware performance modeling engine} that integrates validated single-node predictors, and a~\textit{performance analyzer} that records both static and dynamic system metrics.

The configuration parser ingests system specifications from a YAML file that defines device types (e.g., model, hardware), network links (e.g., RTT, jitter), and runtime policies. An~\texttt{auto\_topology} pass expands this high-level specification into explicit draft and target device pools with fully defined network connections, enabling rapid system configuration and large-scale parameter sweeps.

The core DSD scheduler leverages SimPy to model draft and target servers as concurrent processes, each with explicit queues for batch formation and request scheduling. Network links are represented as delay elements associated with send and receive events, parameterized by configurable RTT and jitter. This request-level abstraction allows fast exploration of large design spaces while preserving accuracy in modeling system-level queueing and network dynamics.

To accurately capture per-kernel execution latency without reimplementing low-level GPU models, DSD-Sim incorporates a hardware performance modeling engine based on VIDUR~\cite{vidur_mlsys2024}. VIDUR provides validated latency predictors for both \emph{prefill} and \emph{decode} operations, grounded in empirical measurements from contemporary GPU architectures.

During simulation, the DSD scheduler communicates with VIDUR’s single-node predictors via a unified API, \texttt{predict(op, shape, hardware)}, which allows querying inference latency for arbitrary batch compositions across heterogeneous devices. This design enables DSD-Sim to retain VIDUR’s single-node accuracy while extending it to distributed inference scenarios. To further adapt VIDUR to distributed speculative decoding workloads, we expand its profiling dataset with measurements from NVIDIA A40 GPUs and edge-oriented LLMs such as Qwen-7B~\cite{Qwen} and Llama2-7B~\cite{Llama}. These additions enable DSD-Sim to model latency behaviors under heterogeneous and resource-constrained environments, improving its predictive accuracy for real-world edge deployments. 

\subsection{Workloads and Trace Model}

DSD-Sim is driven by workload traces that embed request parameters and ground-truth speculation outcomes captured from hardware.

\paragraph{Trace Schema} 
Traces are derived from representative benchmarks spanning diverse LLM tasks such as GSM8K~\cite{cobbe2021gsm8k}, CNN/DailyMail~\cite{hermann2015teaching}, and HumanEval~\cite{chen2021codex}. Each record specifies the parameters needed to drive a simulation run. Crucially, the embedded acceptance sequence is captured from real hardware profiling runs, which faithfully reproduce speculation behavior for a given draft–target pair without relying on probabilistic acceptance models. Table~\ref{tab:trace-fields} lists the key fields in a trace record. The selected benchmarks capture a wide range of LLM inference behaviors, including reasoning-intensive, summarization, and code-generation tasks, covering different output-to-input ratios and speculation acceptance dynamics. This diversity ensures that DSD-Sim operates under representative real-world LLM serving conditions.

\begin{table}[t]
\centering
\caption{Key fields of a workload trace record.}
\label{tab:trace-fields}
\begin{tabular}{@{}ll@{}}
\toprule
Field Name             & Example Value                                \\ \midrule
\texttt{prompt\_length}    & \texttt{27}      \\
\texttt{output\_length}    & \texttt{94}       \\
\texttt{acceptance\_seq}  & \texttt{[1, 0, 1, ...]} \\
\texttt{arrival\_time\_ms}     & \texttt{5.3}            \\
\texttt{drafter\_id}  & \texttt{38}       \\ \bottomrule
\end{tabular}
\end{table}

\paragraph{Arrival Process} Request arrivals can operate in two modes: (i) trace-driven, where timestamps are replayed from a captured workload, and (ii) synthetic, where arrivals follow a Poisson process with a specified rate to emulate stochastic load. In the synthetic mode, arrivals are generated globally and uniformly distributed across drafter devices.

\subsection{Execution Semantics of DSD-Sim}
\label{sec:execution-semantics}
Each request follows a lifecycle that includes routing, batching, and iterative speculation until the generation process completes. DSD-Sim supports two execution modes:
(i) \textbf{Fused mode}, in which both draft and target models are co-located on the same server. The entire speculation loop executes locally, eliminating network overhead between the drafter and verifier.
(ii) \textbf{Distributed mode}, where a draft device proposes a window of tokens (speculation window size $\gamma > 1$), which are then transmitted to a remote target device for parallel verification.

For each arriving request, the simulation progresses through \textbf{Routing, Batching, Speculation,} and \textbf{Verification} stages. During \textit{Routing}, the request is directed to a target cluster according to the active scheduling policy. In \textit{Batching}, it is placed into a queue and grouped with other requests for prefill or decode operations. During \textit{Speculation}, the active $\gamma$ policy determines the window size $\gamma$ and execution mode; in distributed mode, a drafter device generates $\gamma$ speculative tokens and transmits them to the target for verification. In \textit{Verification}, the target server compares these tokens with its own predictions. Accepted tokens advance the generated sequence, while mismatched tokens are corrected using the target’s prediction, and any remaining speculative tokens are discarded. This iterative process continues until the desired output length is reached or an end-of-sequence token is produced.

\subsection{DSD Scheduling Policies}
\label{sec:policies-description}
The execution lifecycle is governed by three families of pluggable policies. Each policy operates on a read-only snapshot of recent system performance metrics, such as queue depth, round-trip time (RTT), time-per-output-token (TPOT), and acceptance rate.

\paragraph{Request Routing Policy} These policies determine how incoming requests are distributed across clusters, supporting a variety of load-balancing strategies in multi-cluster deployments. Common implementations include random selection, round-robin scheduling, and the Join-the-Shortest-Queue (JSQ) policy.

\paragraph{Batching Policy} These policies govern how requests are grouped for prefill and decode operations. They are controlled by configurable parameters for batch size and batching window, and can optionally enable advanced strategies like continuous batching or chunked prefills.

\paragraph{Window Size Policy} These policies dynamically select the appropriate speculation window size ($\gamma$) and execution mode (fused vs. distributed) for each iteration. The simulator provides several such policies: \textit{Static window}, which fixes $\gamma$ to a constant value; \textit{Dynamic window}, which uses threshold-based heuristics to adjust $\gamma$ based on the recent acceptance rate; and our primary learned policy, \textit{Adaptive Window Control} (AWC), which is described in detail in Section~\ref{sec:awc}.

\subsection{Performance Metrics}

To enable policy optimization and offline analysis, DSD-Sim captures detailed performance data for processing by the Performance Analyzer. This data serves as ground truth for evaluating system performance against SLOs and as training input for AWC (Section~\ref{sec:awc}). Specifically, we collect two categories of metrics:

\begin{itemize}
    \item \textbf{Per-Request Metrics:} Time-to-first-token (TTFT), time-per-output-token (TPOT), end-to-end latency, acceptance ratios, routing decisions, and the sequence of window size ($\gamma$) decisions.
    \item \textbf{System-Level Metrics:} Overall system throughput, target device utilization, and aggregate network queueing delays.
\end{itemize}

All measurements are emitted in a structured JSON format to facilitate both online policy adaptation and offline analysis.

\section{Adaptive Window Control}
\label{sec:awc}
The choice of the speculation window size $\gamma$ is crucial for the overall throughput of the distributed framework. If $\gamma$ is set too large, it increases the latency of the draft model and raises the likelihood of verification failures. Conversely, setting $\gamma$ too small results in overly frequent verification steps, significantly increasing both the computational load on the cloud LLM and the communication overhead required for verification tasks.

In real-world deployments, factors such as network latency, token acceptance rate, and device utilization often fluctuate over time, making it challenging for fixed or heuristic-based $\gamma$ policies to consistently balance efficiency and responsiveness. A static speculation window can easily cause excessive rollback overhead under adverse network conditions or lead to severely underutilized hardware when more aggressive speculation would be beneficial. Moreover, it fails to adapt to dynamic system-level conditions in large-scale distributed environments, such as network congestion, queue buildup, and evolving workload characteristics.

To address this issue, Adaptive Window Control (AWC) dynamically adjusts the window size $\gamma$ to achieve superior overall speedup in DSD. By modeling the correlations among system load, communication delay, and token acceptance dynamics, AWC offers a principled, data-driven approach to maintaining high speedup and stable performance across heterogeneous hardware and network conditions.

\subsection{Data-Driven Window Predictor Design}
During runtime, AWC utilizes a window control deep neural network (WC-DNN) that receives input from the performance analyzer. The analyzer encodes both current and historical system states into a feature vector, which the DNN uses to predict the optimal value of $\gamma$. Specifically, this feature vector includes the following information:

\begin{itemize}
    \item \textbf{Queue Depth Utilization ($q_{\text{depth}}$):} Recent utilization rate of the target queue, reflecting the current level of congestion.
    \item \textbf{Acceptance Rate ($\alpha_{\text{recent}}$):} Recent token acceptance ratio from the target model, reflecting the quality and reliability of the draft model’s generated tokens.
    \item \textbf{Per-Link RTT Statistics ($\text{RTT}_{\text{recent}}$):} Recent round-trip latency of the network connection, used to determine the viability of distributed execution.
    \item \textbf{TPOT Statistics ($\text{TPOT}_{\text{recent}}$):} Recent time per output token, representing the processing efficiency of the target model.
    \item \textbf{Prior Speculation Window Size ($\gamma_{\text{prev}}$):} Speculation window size from the previous iteration, providing temporal context for adaptive adjustment.
\end{itemize}
The WC-DNN processes this input to produce the optimal prediction for $\gamma$. The compact size of the input feature vector allows the WC-DNN to operate with minimal computational and memory overhead, thereby ensuring fast and efficient adaptation during runtime.

\subsection{Dataset Generation}
Training the WC-DNN requires labeled data that maps system feature contexts to their optimal window size configurations. To create this dataset, we perform exhaustive sweeps of the window size under various system conditions. For each experimental setup, defined by a workload trace, network configuration, and hardware deployment, the simulator runs across all combinations of window sizes (from 2 to 12) as well as the fused execution mode.

During each sweep, the system records a set of data that includes the feature vector, policy output, and performance metrics such as TTFT and TPOT. After gathering results from more than 2,000 scenarios, we construct training labels by selecting the configuration that minimizes a weighted objective function reflecting the SLO priorities.

\subsection{Model Architecture and Training}
\begin{figure}[t]
  \centering
  \includegraphics[width=\columnwidth]{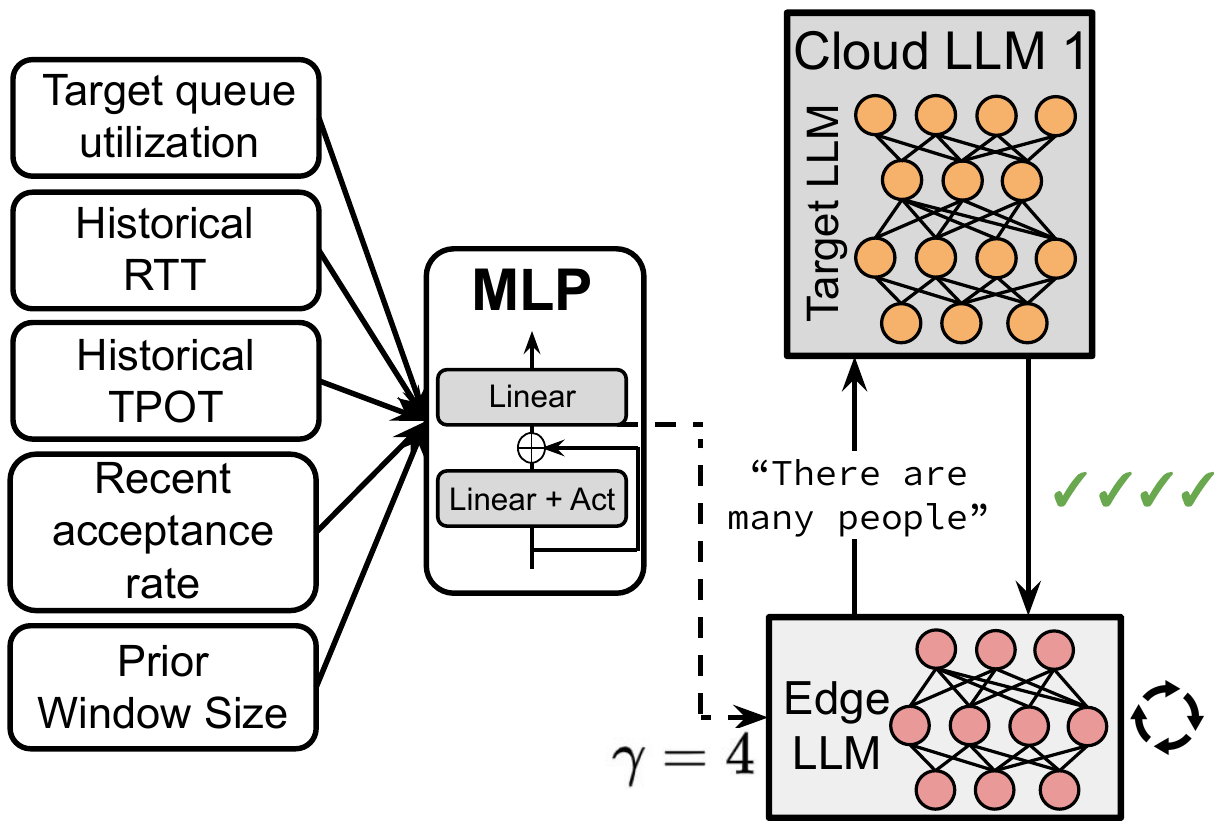}
  \caption{The WC-DNN architecture takes five input features and generates the optimal prediction for the speculation window size. The edge LLM then adjusts its window size based on this prediction and coordinates with the cloud LLM for verification.}
  \label{fig:awc}
\end{figure}

The WC-DNN adopts a residual multi-layer perceptron (MLP) architecture that predicts the speculation window size $\gamma$ as a continuous value, as shown in Figure~\ref{fig:awc}. The MLP takes a five-dimensional feature vector as input and employs a residual structure with two computational blocks and SiLU activation functions to generate a stable scalar prediction for the window size. The network is trained using supervised regression with an $L_1$ loss, and we use the AdamW optimizer for 100 epochs, achieving consistently high predictive accuracy on the validation set.

\subsection{Stable Execution of Window Size Prediction}
In practice, directly executing the trained WC-DNN can lead to fluctuations in the predicted window size $\gamma$ because of variations in system metrics, which in turn cause instability in overall system performance. To address this issue, we introduce three stabilization techniques:
\begin{enumerate}
    \item \textbf{Clamping:} Window size predictions are clipped to a configured range (e.g., $[1, 12]$).
    \item \textbf{Exponential Smoothing:} An Exponential Moving Average (EMA) with a smoothing factor $\alpha = 0.4$ is applied to the predicted window size across iterations to dampen high-frequency oscillations.
    \item \textbf{Hysteresis for Mode Switching:} A ``sticky'' policy prevents rapid toggling between fused and distributed modes. When the system is in distributed mode, the smoothed prediction must remain near $\gamma=1$ for at least $k$ consecutive steps (typically $k=2$) before a switch to fused mode is permitted.
\end{enumerate}

After applying the stabilization techniques, the predicted window size is quantized to the nearest integer within the predefined range (e.g., $[1,12]$) and then used in the next iteration of SD. The smoothing state is maintained per draft–target pair so each connection follows its own trajectory, but shared features ensure that decisions still reflect aggregate system conditions.

Crucially, if the window size predicted by the WC-DNN is less than or equal to one ($\gamma \le 1$), the system switches to \textbf{Fused Mode}. In this state, the cloud LLM generates all tokens directly, bypassing the draft model. This situation typically arises when the edge device operates very slowly or when network conditions are severely congested.

\section{Evaluation}
We evaluate DSD-Sim across two primary dimensions: (1) \emph{System Performance and Scalability}, by testing diverse configurations and workload compositions across heterogeneous Edge–Cloud setups; and (2) \emph{Policy Effectiveness}, by analyzing how the proposed AWC strategy performs relative to conventional scheduling and batching baselines.

\begin{itemize}
\item \textbf{Calibration:} How closely do VIDUR’s latency predictions and our RTT assumptions match real hardware measurements?
\item \textbf{Scalability:} How does distributed speculative decoding behave as the system scales in heterogeneity and the number of participating nodes?
\item \textbf{Policy Efficiency:} How effectively does AWC enhance resource utilization and reduce the overall LLM execution latency compared to other baseline policies?
\end{itemize}

All experiments are conducted using workload traces derived from GSM8K~\cite{cobbe2021gsm8k}, CNN/DailyMail~\cite{hermann2015teaching}, and HumanEval~\cite{chen2021codex} benchmarks. Specifically, we use Qwen-7B~\cite{Qwen} and Llama2-7B~\cite{Llama} running on NVIDIA A40 GPUs~\cite{nvidia_a40} to model edge-side inference, while Llama2-70B~\cite{Llama} and Qwen-72B~\cite{qwen72B} are deployed on A100~\cite{nvidia_a100} and H100 GPUs~\cite{nvidia_h100}, respectively, to represent cloud-scale execution.
Each measurement is repeated across multiple random seeds, and the reported results represent the mean values to reduce the influence of stochastic variations. 

\begin{figure}[t]
  \centering
  \includegraphics[width=\columnwidth]{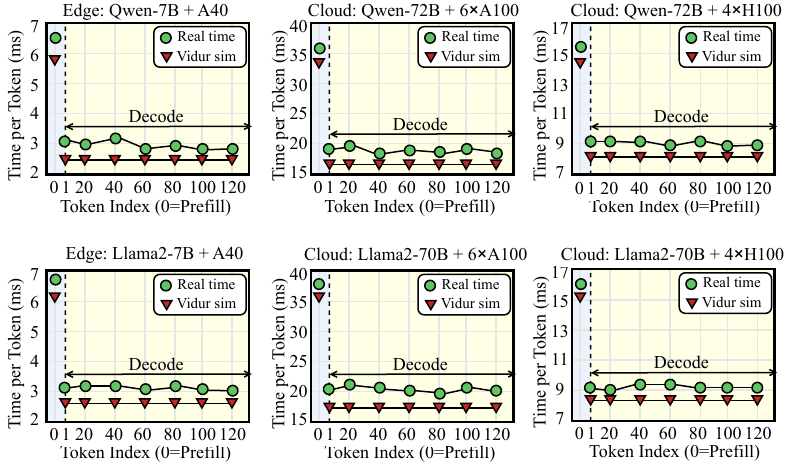}
  \caption{GPU-level calibration of predicted vs. actual inference latencies for prefill and decode across Qwen-7B, Qwen-72B, Llama-2-7B, and Llama-2-70B on A40, A100, and H100 GPUs. Error bars indicate standard deviation over 100 requests.}
  \label{fig:gpu-calibration}
\end{figure}

\subsection{Calibration and Validation}
\label{sec:fidelity-eval}
DSD-Sim is composed of several key components: a configuration parser that loads traces and system settings, a deterministic discrete-event scheduler that applies the specified routing, queueing, and window policies, a hardware modeling engine that estimates GPU inference latency, and a performance analyzer that records and reports system-level metrics.
Only the hardware modeling engine, which is built on VIDUR, and the assumed edge–cloud RTT introduce modeling uncertainty; the parser and scheduler simply replay traces and apply policies without any tunable parameters. Therefore, we validate VIDUR’s prefill and decode latency predictions against real hardware measurements and determine RTT values using real measurements from prior latency studies. With these components calibrated, the simulator can reasonably estimates system behavior while maintaining a controlled and reproducible evaluation setting.
\begin{figure*}[t]
  \centering
  \includegraphics[width=\textwidth]{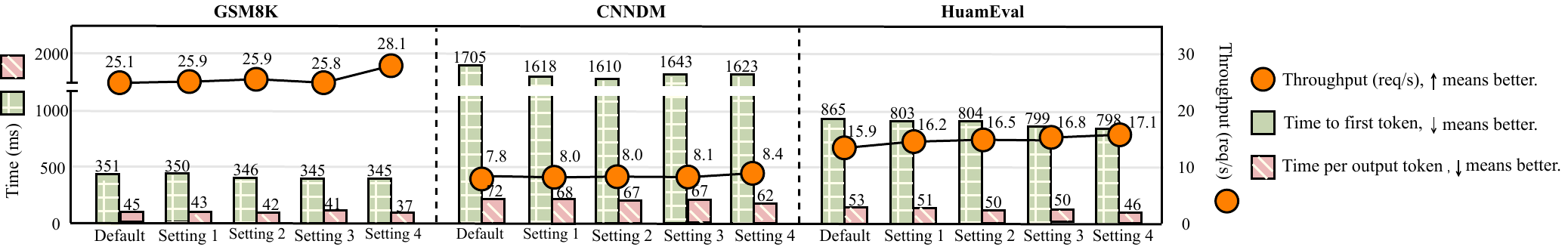}
  \caption{End-to-end SLOs and throughput for policy stacks. Default: Random routing + FIFO queueing + Static $\gamma$. Setting~1: JSQ + FIFO + Static $\gamma$. Setting~2: JSQ + Length-Aware Batching (LAB) + Static $\gamma$. Setting~3: JSQ + Length-Aware Batching + Dynamic $\gamma$. Setting~4: JSQ + Length-Aware Batching + AWC.}
  \label{fig:system-configs}
\end{figure*}
\paragraph{GPU-Level Calibration}
Figure~\ref{fig:gpu-calibration} presents a comparison between DSD-Sim’s predicted inference latencies and empirical measurements obtained from real GPU deployments across different model sizes and batch configurations. In particular, we evaluate simulation fidelity using multiple large language models, including Qwen-7B, Qwen-72B~\cite{team2024qwen2}, Llama2-7B, and Llama2-70B~\cite{touvron2023llama}, executed on diverse real GPU platforms such as NVIDIA A40, A100, and H100. To ensure consistency in workload patterns, all models are benchmarked on prompts derived from the GSM8K dataset, which provides realistic reasoning-oriented text sequences representative of modern LLM inference workloads. In parallel, the hardware modeling engine is executed to estimate the corresponding simulated latencies, enabling direct calibration and cross-validation against real-world hardware performance Specifically, we adopt \textit{VIDUR}~\cite{vidur_mlsys2024} to simulate GPU hardware performance. For smaller models such as Qwen-7B and Llama2-7B, a single GPU is sufficient to accommodate the entire model. In contrast, larger models such as Qwen-72B and Llama2-70B require multi-GPU execution, where model parallelism is employed to distribute weights and computation across multiple devices.

The simulator integrates \textit{VIDUR}’s performance predictors and achieves a mean absolute error of $7.4\%$ for prefill latency and $5.2\%$ for decode latency across all evaluated configurations. This level of accuracy validates the simulator’s reliability in modeling single-node LLM inference behavior.

It is worth noting that \textit{VIDUR}’s simulated latencies are consistently lower than empirical measurements across all model and GPU setups. This systematic deviation stems from the fact that \textit{VIDUR} focuses exclusively on modeling MLP and Attention kernel execution times, omitting NCCL communication overheads~\cite{nccl_link} and other non-kernel computations that occur in real hardware environments. 

\paragraph{Network Calibration}
We model the edge–to–cloud communication link using two representative round-trip time (RTT) settings. According to public Azure latency reports, the RTT to nearby East-US regions is typically below 20~ms~\cite{azure-latency-stats}, while independent measurement studies on Azure and AWS observe values ranging from 10 to 30~ms for clients located in the same geographic area as the serving region~\cite{palumbo2020cloudlatency}. Guided by these observations, we evaluate network performance under two conditions—10 ms representing the typical case and 30 ms representing the upper bound of common operating scenarios. DSD-Sim reports performance metrics for both networking settings.

\subsection{System Performance Analysis}
In this section, we evaluate the capacity of DSD-Sim to model distributed speculative decoding across a range of hardware configurations, network conditions, and execution modes.

\paragraph{Performance Across System Configurations}
We define a large-scale heterogeneous cluster comprising a \textbf{Cloud Pool} and an \textbf{Edge Pool}. The Cloud Pool consists of 20 servers hosting large models (LLaMA2-70B, LLaMA3-70B, Qwen-72B) distributed across 4$\times$A100, 4$\times$H100, and 4$\times$A6000 configurations. The Edge Pool contains 600 GPUs (300 A40s and 300 V100s) evenly serving LLaMA-2-7B, Qwen-7B, and LLaMA-3.1-8B draft models. All experiments run over a 10~ms RTT network connection.

The scheduler exposes three knobs: \textbf{Routing} (Random vs.\ JSQ); \textbf{Queueing} (FIFO vs.\ Length-Aware Batching, or LAB, which groups requests of similar lengths to minimize padding); and \textbf{Window Control} (Static $\gamma=4$, Dynamic $\gamma$, and AWC).
Figure~\ref{fig:system-configs} compares five policy stacks: Default (Random + FIFO + Static $\gamma$), Setting~1 (JSQ + FIFO + Static $\gamma$), Setting~2 (JSQ + LAB + Static $\gamma$), Setting~3 (JSQ + LAB + Dynamic $\gamma$), and Setting~4 (JSQ + LAB + AWC).

Accumulating these policies yields steady improvements in both throughput and latency. Across GSM8K, for example, throughput climbs from 25.1 to 28.1 requests/s as we add all four components, while TTFT drops from 351 to 345\,ms and TPOT from 45 to 37\,ms. CNN/DailyMail and HumanEval exhibit similar trends, with AWC providing the major latency gain. The overall results demonstrate that coordinating routing, batching, and adaptive window control jointly offers the optimal throughput–latency trade-off.

\begin{figure}[t]
  \centering
  \includegraphics[width=\columnwidth]{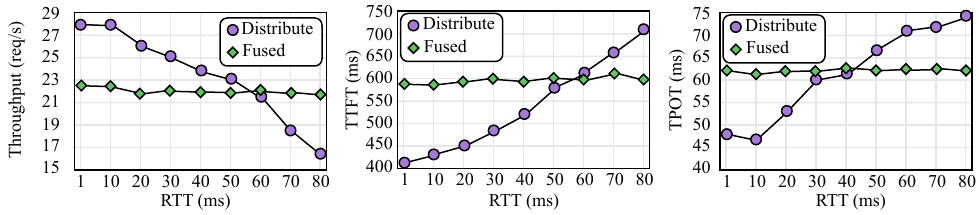}
  \caption{Throughput, TTFT, and TPOT for distributed (purple) vs.\ fused (green) execution as RTT increases. Distributed excels at low RTT but degrades once network delay dominates; fused remains steady because work stays local.}
  \label{fig:fused-vs-distributed}
\end{figure}

\paragraph{Distributed vs.\ Cloud-Only Execution}
We next analyze how network RTT affects the performance of distributed speculative decoding. Figure~\ref{fig:fused-vs-distributed} varies RTT while keeping all other system configurations fixed. At low RTT values, distributed execution outperforms cloud-only mode because edge drafts are generated concurrently with cloud verification, improving throughput. However, as RTT increases, the communication overhead of each speculative iteration grows, causing noticeable degradation. In contrast, cloud-only mode (where the target handles both drafting and verification locally) remains unaffected by network latency but suffers from lower parallelism at low RTTs. The performance crossover observed around 50–60\,ms explicitly highlights the trade-off between compute offloading and network overhead, validating the need for AWC's adaptive mode switching.

\paragraph{AWC vs.\ Baseline Policies}

We next evaluate AWC against two baselines: \emph{Static Window} (fixed $\gamma=4$) and \emph{Dynamic Window} (increments $\gamma$ when the recent acceptance rate exceeds $0.75$, and decrements when it falls below $0.25$). Table~\ref{tab:awc-comparison} presents results across four system configurations, combining 20 targets with either 600 or 1000 drafts and network RTTs of 10~ms or 30~ms, evaluated on three datasets: GSM8K, CNN/DailyMail, and HumanEval. AWC delivers the best throughput in all 12 cases, improving over the static baseline by 3--10\% (e.g., +9.7\% on GSM8K with 600 drafts and +4.1\% on CNNDM with 1000 drafts) and outperforming the dynamic baseline in 11 out of 12 scenarios.

Latency metrics closely mirror these overall trends. TTFT remains within 0.5--4\% of the best baseline, and TPOT consistently drops by 6--10\% across the evaluated workload set (400 GSM8K, 400 CNN/DailyMail, and 100 HumanEval prompts). These results clearly show that a learned controller can simultaneously raise throughput and lower latency across heterogeneous conditions without manual threshold tuning.

\begin{table*}[t]
  \centering
  \caption{Adaptive window control versus baseline $\gamma$ policies. Values are averaged over three runs; parentheses indicate improvement relative to the Static baseline (positive for higher throughput, negative for lower latency).}
  \label{tab:awc-comparison}
  \resizebox{\textwidth}{!}{
  \begin{tabular}{lcccccccccccc}
    \toprule
    & \multicolumn{3}{c}{Config 1: 20T/600D, 10ms RTT} &
      \multicolumn{3}{c}{Config 2: 20T/1000D, 10ms RTT} &
      \multicolumn{3}{c}{Config 3: 20T/600D, 30ms RTT} &
      \multicolumn{3}{c}{Config 4: 20T/1000D, 30ms RTT} \\
    \cmidrule(lr){2-4}\cmidrule(lr){5-7}\cmidrule(lr){8-10}\cmidrule(lr){11-13}
    & Static & Simple & AWC &
      Static & Simple & AWC &
      Static & Simple & AWC &
      Static & Simple & AWC \\
    \midrule
    \multicolumn{13}{l}{\textbf{Throughput (requests/s) $\uparrow$}} \\
    GSM8K & 25.8 & 26.1 & 28.3 ($+9.7\%$) &
            30.6 & 30.7 & 31.5 ($+3.0\%$) &
            17.8 & 18.9 & 18.5 ($+3.9\%$) &
            21.4 & 21.8 & 22.3 ($+4.2\%$) \\
    HumanEval & 16.2 & 16.7 & 17.2 ($+6.1\%$) &
                17.4 & 17.6 & 18.2 ($+4.6\%$) &
                11.2 & 11.0 & 11.3 ($+0.9\%$) &
                12.8 & 13.1 & 13.2 ($+3.1\%$) \\
    CNNDM & 8.0 & 8.25 & 8.4 ($+5.0\%$) &
            12.1 & 12.2 & 12.6 ($+4.1\%$) &
            6.0 & 6.2 & 6.2 ($+3.3\%$) &
            9.3 & 9.5 & 9.7 ($+4.3\%$) \\
    \midrule
    \multicolumn{13}{l}{\textbf{TTFT (ms) $\downarrow$}} \\
    GSM8K & 352 & 334 & 347 ($-1.4\%$) &
            368 & 354 & 362 ($-1.6\%$) &
            462 & 457 & 443 ($-4.1\%$) &
            478 & 465 & 463 ($-3.1\%$) \\
    HumanEval & 803 & 792 & 799 ($-0.5\%$) &
                2102 & 2178 & 2038 ($-3.0\%$) &
                1138 & 1125 & 1120 ($-1.6\%$) &
                2578 & 2520 & 2497 ($-3.1\%$) \\
    CNNDM & 1602 & 1625 & 1569 ($-2.1\%$) &
            2412 & 2408 & 2352 ($-2.5\%$) &
            2024 & 1993 & 2006 ($-0.9\%$) &
            2986 & 2934 & 2925 ($-2.0\%$) \\
    \midrule
    \multicolumn{13}{l}{\textbf{TPOT (ms) $\downarrow$}} \\
    GSM8K & 40.1 & 38.9 & 36.2 ($-10.7\%$) &
            72.2 & 69.7 & 67.4 ($-7.1\%$) &
            62.3 & 62.9 & 58.3 ($-6.8\%$) &
            99.4 & 98.1 & 91.5 ($-8.6\%$) \\
    HumanEval & 50.8 & 48.6 & 46.3 ($-9.7\%$) &
                75.4 & 73.4 & 70.5 ($-6.9\%$) &
                79.2 & 78.1 & 73.6 ($-7.0\%$) &
                102.7 & 100.2 & 96.8 ($-6.0\%$) \\
    CNNDM & 67.4 & 66.9 & 62.8 ($-6.8\%$) &
            82.1 & 82.6 & 76.8 ($-6.5\%$) &
            92.4 & 88.6 & 86.7 ($-6.2\%$) &
            108.3 & 108.5 & 98.2 ($-10.2\%$) \\
    \bottomrule
  \end{tabular}}
\end{table*}

\subsection{Ablation Studies}
\label{sec:ablations}
In this section, we perform systematic ablation studies to isolate and quantify the effects of key design choices in scheduling, queueing, and batching policies. While the previous section demonstrated the aggregate benefits of the full DSD stack, identifying the individual contribution of each component is critical for system optimization. The complex interaction between load balancing, batch formation, and window sizing often obscures the specific source of performance gains. Therefore, we decouple these mechanisms to analyze how routing decisions impact server utilization and how batching strategies mitigate head-of-line blocking.

\begin{figure}[t]
  \centering
  \includegraphics[width=\columnwidth]{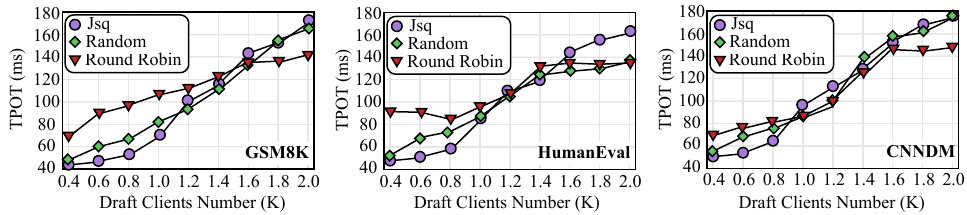}
  \caption{TPOT versus number of draft clients for three routing policies (GSM8K, HumanEval, CNNDM).} 
  \label{fig:routing-ablation}
\end{figure}

\paragraph{Request Routing Strategies}
Figure~\ref{fig:routing-ablation} evaluates system performance by varying the number of draft models from 0.4k to 2.0k while comparing three routing strategies: Random, Round-Robin (RR), and Join-Shortest-Queue (JSQ), as described in Section~\ref{sec:policies-description}.

JSQ achieves the best performance when system resources are not fully utilized, maintaining TPOT values 5–20 ms lower across all traces by routing requests to less busy target servers. However, as the system approaches its maximum capacity, the fastest server remains saturated, causing subsequent requests already assigned to it to suffer from head-of-line blocking. As a result, the TPOT curve eventually rises above that of Round-Robin.

\begin{figure}[t]
  \centering
  \includegraphics[width=\columnwidth]{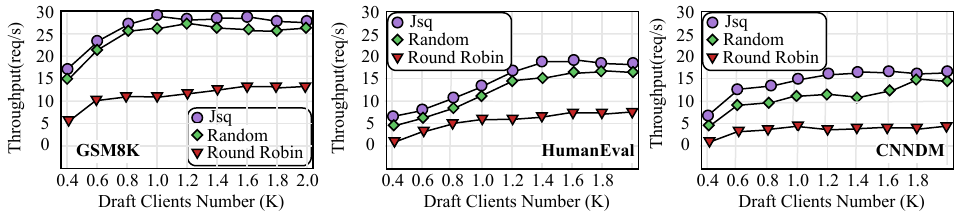}
  \caption{Throughput vs. the number of draft clients for the same routing policies. JSQ consistently delivers the best throughput up to \(\sim\)1\,k drafts, but gradually saturates thereafter.}
  \label{fig:routing-throughput}
\end{figure}

Although RR is not load-aware, it distributes user requests more evenly across servers. The throughput trends in Figure~\ref{fig:routing-throughput} reflect a similar pattern: JSQ delivers the highest scalability up to around one thousand draft models but then plateaus as resources become saturated, while RR continues to improve and eventually catches up. These results indicate that adaptive scheduling strategies, which respond to changing system conditions, can outperform static policies and maintain stable performance across varying load levels.

\begin{figure}[t]
  \centering
  \includegraphics[width=\columnwidth]{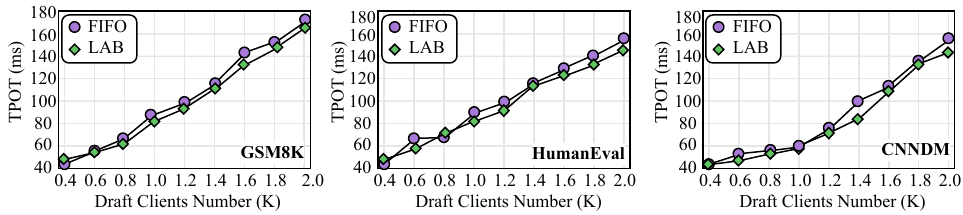}
  \caption{FIFO versus length-aware batching (LAB) performance comparison across diverse workloads. LAB groups similar-length jobs, improving TPOT under moderate-to-high load conditions.}
  \label{fig:throughput-queueing-ablation}
\end{figure}

\begin{figure}[t]
  \centering
  \includegraphics[width=\columnwidth]{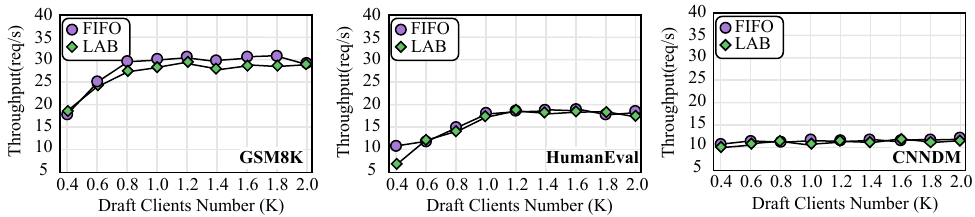}
  \caption{FIFO versus length-aware batching (LAB) across workloads. LAB groups similar-length jobs, improving throughput.}
  \label{fig:queueing-ablation}
\end{figure}

\paragraph{Queueing and Batching Policies}
We compare simple FIFO dispatch with a length-aware batching (LAB) policy. LAB takes the head-of-line request and batches it with other requests whose lengths closely match the head-of-line request. It is widely adopted (e.g., ORCA~\cite{yu2022orca}, Sarathi~\cite{agrawal2023sarathi}) and serves as the strongest length-aware baseline. Figure~\ref{fig:throughput-queueing-ablation} illustrates the latency benefit of this strategy. LAB consistently achieves lower TPOT compared to FIFO (reducing latency by 1--2\,ms) across all workloads. By grouping requests with similar execution lengths, LAB minimizes the padding overhead within a batch, effectively mitigating head-of-line blocking where short requests wait for long ones. In terms of scalability, Figure~\ref{fig:queueing-ablation} demonstrates that both policies reach similar throughput ceiling. As the number of draft clients increases beyond 1k, the system saturates, and throughput plateaus for both FIFO and LAB. This confirms that while LAB optimizes execution efficiency (improving latency), the maximum aggregate throughput is determined by the underlying compute capacity rather than the queueing order.

\section{Conclusion}
We presented DSD, a distributed speculative decoding
framework for efficient large model serving across
edge–cloud environments. Built upon the DSD-Sim simulator,
DSD models realistic network, batching, and scheduling
dynamics to guide performance optimization. We further
introduced AWC, a data-driven policy that dynamically adjusts
the speculation window for optimal throughput and
stability. Experiments show that DSD greatly improves
latency and scalability over existing SD methods.

\newpage
\bibliography{example_paper}
\bibliographystyle{mlsys2025}


\end{document}